\documentclass[10pt,twocolumn,letterpaper]{article}

\usepackage[pagenumbers]{cvpr}              
\usepackage{placeins}
\usepackage{placeins}
\usepackage{booktabs}
\usepackage{multirow}
\usepackage{adjustbox}
\usepackage{amsmath}
\usepackage{float}
\usepackage{comment}
\usepackage{algorithm}
\usepackage{algpseudocode}
\usepackage{tocloft}

\newcommand{\kron}{\otimes}

\newcommand{\bB}{\mathbf{B}}
\newcommand{\bC}{\mathbf{C}}

\newcommand{\bH}{\mathbf{H}}
\newcommand{\bI}{\mathbf{I}}

\newcommand{\bK}{\mathbf{K}}

\newcommand{\bP}{\mathbf{P}}

\newcommand{\bX}{\mathbf{X}}
\newcommand{\bY}{\mathbf{Y}}

\newcommand{\bu}{\mathbf{u}}
\newcommand{\bv}{\mathbf{v}}

\newcommand{\bx}{\mathbf{x}}
\newcommand{\by}{\mathbf{y}}
\newcommand{\bz}{\mathbf{z}}

\definecolor{cvprblue}{rgb}{0.21,0.49,0.74}
\usepackage[pagebackref,breaklinks,colorlinks,allcolors=cvprblue]{hyperref}

\def\paperID{****}
\def\confName{****}
\def\confYear{2026}

\title{Spectrum from Defocus: Fast Spectral Imaging with Chromatic Focal Stack}

\author{
    M. Kerem Aydin\textsuperscript{1,*} \quad
    Yi-Chun Hung\textsuperscript{1,*} \quad
    Jaclyn Pytlarz\textsuperscript{2} \quad
    Qi Guo\textsuperscript{3} \quad
    Emma Alexander\textsuperscript{1} \\
    \textsuperscript{1}Department of Computer Science, McCormick School of Engineering, Northwestern University \\
    \textsuperscript{2}Dolby Laboratories, Inc. 
    \textsuperscript{3}Elmore Family School of Electrical and Computer Engineering, Purdue University \\
    \textsuperscript{*} Equal contribution
}

\begin{document}
\maketitle

\begin{abstract}
Hyperspectral cameras face harsh trade-offs between spatial, spectral, and temporal resolution in inherently low-photon conditions. Computational imaging systems break through these trade-offs with compressive sensing, but have required complex optics and/or extensive compute.
We present Spectrum from Defocus (SfD), a chromatic focal sweep method that achieves state-of-the-art hyperspectral imaging with only two off-the-shelf lenses, a grayscale sensor, and less than one second of reconstruction time.
By capturing a chromatically-aberrated focal stack that preserves nearly all incident light, and reconstructing it with a fast physics-based iterative algorithm, SfD delivers sharp, accurate hyperspectral images.
The combination of photon efficiency, optical simplicity, and physical interpretability makes SfD a promising solution for fast, compact, interpretable hyperspectral imaging.
\end{abstract}
    
\section{Introduction}
\label{sec:intro}

\begin{figure*}[ht!]
    \centering
    \includegraphics[width=1\linewidth]{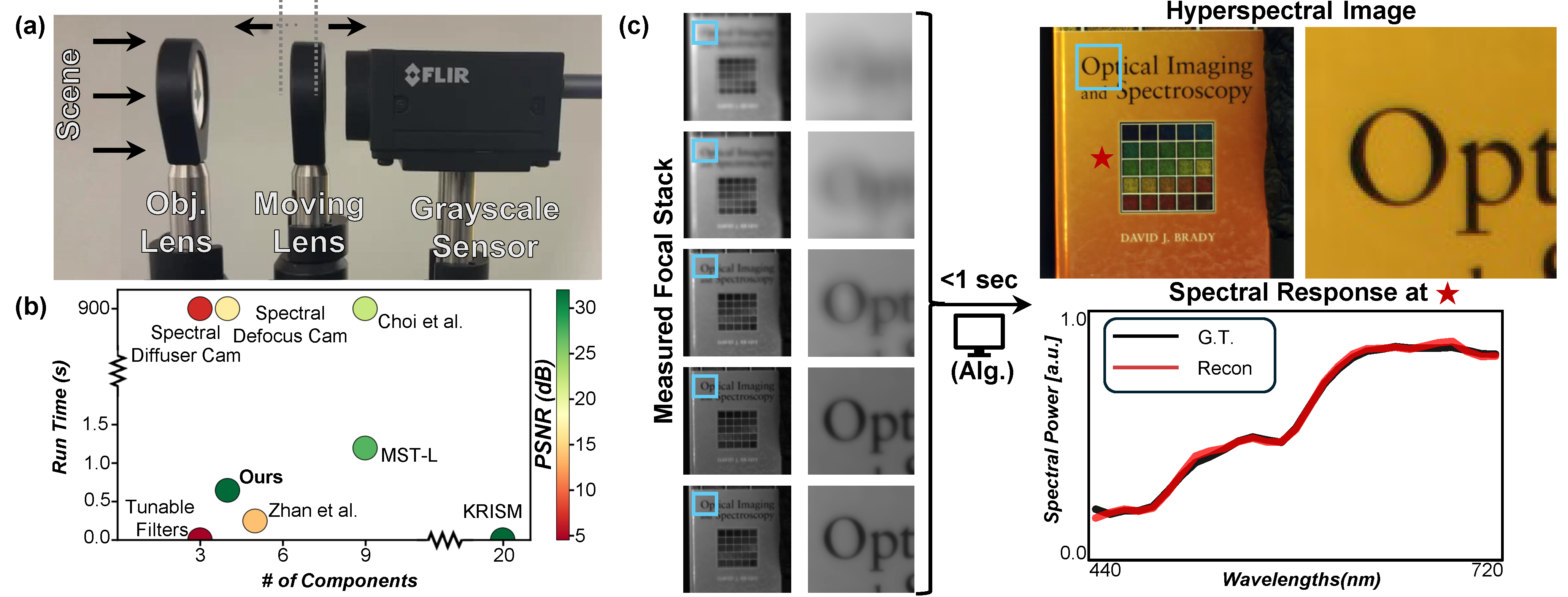}
    \caption{\textbf{The Spectrum from Defocus (SfD) method.}
    (a) Our hardware prototype uses a moving lens to sweep focus through chromatic aberration. 
    (b) SfD achieves state-of-the-art hyperspectral imaging with simple optics and low computational cost. See Table~\ref{tab:comparison} for details.
    (c) The system captures 5 defocused grayscale images and reconstructs hyperspectral images in under a second.}
    \label{fig:teaser}
\end{figure*}

Hyperspectral imaging (HSI) extends traditional color imaging by capturing detailed spectral signatures across numerous spectral bands, rather than just the three broad RGB bands.
 This richer spectral information is essential for applications that rely on subtle material differences, such as remote sensing for environmental monitoring \cite{bioucas2013hyperspectral}, medical diagnostics \cite{lu2014medical}, food quality control \cite{feng2012application}, and industrial inspection \cite{peyghambari2021hyperspectral}.
Despite its importance, capturing high-quality spectral images with compact, affordable cameras remains challenging.
Hyperspectral sensors suffer from inherently low signal-to-noise ratios because incoming light is divided across many spectral channels, leaving each channel with reduced photon counts.
This challenge is amplified in systems that rely on spectral filters, dispersive elements, or coded apertures, all of which block and/or spread light and further reduce photon efficiency.
As a result, traditional hyperspectral cameras are particularly fragile in low-light conditions or dynamic scenes that require short exposure times, limiting their practicality. 

To mitigate this fragility, compressive spectral imaging systems encode spectral information into spatial measurements, enabling snapshot capture or reduced acquisition times \cite{bacca2023computational}.
These systems typically rely on coded optics followed by computational reconstruction. 
While effective, they often demand heavy computation, involving large matrix operations or iterative solvers. 
Faster alternatives, such as inverse filtering, degrade reconstruction quality and limit their practical use in real-world scenes \cite{zhao2019spectral}.

To address the computational challenges of classical reconstruction methods, learning-based approaches have been increasingly explored.
These methods often incorporate spectral priors into parameterized architectures to improve efficiency and accuracy.
However, these data-driven models risk hallucinating spectral content \cite{fu2024limitations}, making them less reliable in tasks where spectral accuracy is critical, such as material analysis or food quality inspection.

To address these limitations in light inefficiency, high computational complexity, and hallucination, we propose a chromatic focal sweep camera consisting of two lenses and a grayscale sensor.  
By exploiting chromatic aberration as a natural cue for separating wavelengths, our optical system produces a structured sensing matrix that enables computationally efficient and robust hyperspectral reconstruction.
Furthermore, this physics-based approach, though boosted by a deep denoising network, mitigates the hallucination issues commonly observed in purely-data-driven methods.  

The main contributions of this paper are as follows:  
\begin{itemize}  
    \item State-of-the-art spectral image quality through a combination of light-efficient optics and a robust algorithm.  
    \item Fast computation with an interpretable model that leverages both physics- and data-driven components.  
    \item A simple optical design consisting of few, off-the-shelf components, which outperforms systems with up to 20 optical elements.  
\end{itemize}

\section{Related Works}
\label{sec:related_works}

\textit{Scanning Methods} acquire hyperspectral data cubes by sequentially capturing individual 2D slices of the scene. 
The most straightforward implementation uses band-pass filters or liquid crystal tunable filters to capture one narrow spectral band per frame, gradually scanning across the spectrum  \cite{slawson1999hyperspectral, wang2014multispectral, zhang2021deeply}. 
Pushbroom systems take a slightly different approach, capturing a single spatial row at a time through a narrow slit, while dispersive optics spread each row into a full spectrum across the sensor \cite{porter1987system}. 
While these techniques can deliver high spatial and spectral resolution, they suffer from low photon efficiency—since only a fraction of the incoming light is used for each measurement—and long acquisition times, which restricts them to static, well-lit scenes.
In contrast, our method passively encodes spectral information into each frame and avoids light-blocking (low light transmission) elements entirely, enabling faster capture with higher photon throughput.

\textit{Compressive hyperspectral imaging systems} capture optically coded measurements that must be computationally reconstructed to recover the final hyperspectral data cube.
Among these, Coded Aperture Snapshot Spectral Imaging (CASSI) \cite{wagadarikar2008single} was developed to overcome the limitations of sequential scanning systems. 
A coded aperture is placed either before or after a dispersive prism, encoding spectral signatures into the spatial domain. 
CASSI systems are typically divided into two categories based on their encoding strategy: spatially-encoded CASSI, which employs a single disperser (SD-CASSI) \cite{wagadarikar2008single}, and spatial-spectral CASSI, which codes information in both spatial and spectral domains (DD-CASSI \cite{gehm2007single}, SS-CASSI \cite{lin2014spatial}). 
Regardless of the encoding scheme, in all CASSI systems, reconstruction algorithms play a critical role in final image quality. Early approaches relied on optimization-based solvers with hand-crafted priors \cite{wagadarikar2008single, bioucas2007new, liu2018rank, yuan2016generalized}, while more recent approaches introduced learning-based reconstructions \cite{cai2022mask,ma2019deep,cai2022coarse, meng2020end, cheng2022recurrent, hu2022hdnet}. 
Hybrid methods further combined the system’s sensing matrix with deep learning, either to directly parameterize the solution space \cite{choi2017high, ma2019deep, meng2020gap} or to learn data-driven priors for regularization \cite{xie2023plug, yuan2020plug, zheng2021deep}. 
We compare our method to two CASSI systems: Choi et al.~\cite{choi2017high} which uses a hybrid physics- and data-driven reconstruction, 
and the transformer-based MST~\cite{cai2022mask} which uses spectral-wise self-attention to capture correlations across wavelengths.
Despite their success, traditional CASSI systems are often relatively large and mechanically sensitive, making them impractical for compact imaging platforms. In comparison, our system uses only two off-the-shelf lenses and achieves spectral imaging through a focal sweep, significantly reducing hardware complexity.

\textit{Diffractive optical elements (DOEs) and metasurfaces}

have been designed in tandem with computational reconstruction for compressive spectral sensing \cite{habel2012practical, heide2016encoded, shi2024split, hu2024multishot, liu2025metah2, kar2019compressive}. 
Early work, such as~\cite{habel2012practical}, used simple diffraction gratings to simultaneously resolve spatial and spectral information, but these systems suffered from low resolution and bulky optical assemblies. 
More recently, efforts have focused on using a single DOE, reducing hardware complexity but introducing stronger optical artifacts which are typically handled through end-to-end deep learning reconstructions trained on large spectral datasets~\cite{baek2021single, jeon2019compact, li2022quantization}.
We compare our work with a recent approach, 2in1 Cameras~\cite{shi2024split}, which leverages dual pixels to simultaneously image through a DOE and a conventional lens.
Similar to these approaches, our system also relies on spectrally encoded PSFs; however, we achieve this encoding using the naturally occurring chromatic aberration of standard refractive lenses, avoiding the need for custom-fabricated optics.

\begin{figure*}[ht!]
  \centering
  \includegraphics[width=\textwidth]{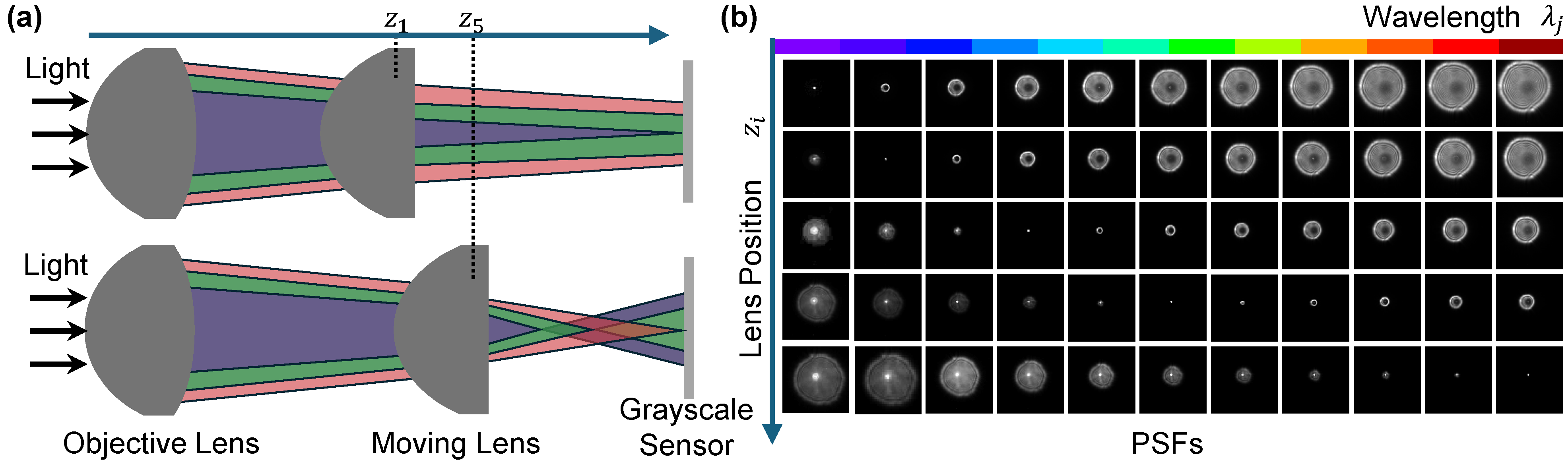}
   \caption{\textbf{Optical design}.
   (a) The system consists of a lens pair in which the second lens is translated to five discrete positions, $z_1, z_2, \ldots, z_5$. At each position $z_i$, the captured image $I_i$ is focused on a corresponding wavelength $\lambda_i$, while other wavelengths appear blurred. Together, these measurements form the chromatic focal stack.
   (b) Measured point spread functions (PSFs) from the real-world prototype at different wavelengths $\lambda_i$ and lens positions $z_i$, showing how the focal plane shifts.  While calibration PSFs are collected with the help of a narrow-band tunable filter, at deployment time our grayscale sensor measures only the scene-weighted sum of incoming light across all wavelengths at each of the five lens positions.
   }
   \label{fig:system}
\end{figure*}

\textit{Other Recent Spectral Computational Imagers} highlight the trade-off between optical simplicity and computational cost in spectral imaging. 
Spectral DiffuserCam (SDC)~\cite{monakhova2020spectral} eliminates conventional lenses by using a diffuser and a specialized sensor equipped with a spectral filter array, analogous to a Bayer pattern. 
By spreading fine spatial details across multiple pixels, SDC encodes high-frequency information that is later recovered via inverse reconstruction with FISTA.
A more recent system, Spectral DefocusCam~\cite{foley2025spectral}, replaces the diffuser with an achromatic lens, enabling faster acquisition of multiple, less blurry measurements, leading to significantly improved reconstruction quality.
Our work shares similarities with Spectral DefocusCam in that both systems encode information via defocus. However, while DefocusCam uses binned super-pixels and exploits blur for spatial super-resolution, our system employs a grayscale sensor and encodes spectral content through spatially varying blur induced by chromatic aberration.
Another direction is represented by KRISM~\cite{saragadam2019krism}, which uses a complex optical design to directly measure spectral singular values, thereby shifting most of the computational burden to optical hardware.
Our approach strikes a balance between these extremes by combining a simple optical system—using only off-the-shelf parts—with a lightweight, physics-based reconstruction algorithm.

Focal sweeping and stacking form the basis of our compressive sensing approach, in which wavelength-dependent defocus encodes spectral information across a series of focal planes.
Variants of this strategy have been explored to extend depth of field \cite{cossairt2010spectral}, enable depth sensing through color-coded apertures \cite{chakrabarti2012depth}, reduce chromatic fringing in simple lenses \cite{heide2013high}, and light field imaging \cite{huang2017multispectral}. Spectral reconstruction from focal stacks has been explored in \cite{he2019spectral}, which recovers spatially sparse spectra from PSF center attenuation, and by Zhan et al.~\cite{zhan2019hyperspectral} to reproduce full hyperspectral images with a naive algorithm. Our robust algorithm produces full hyperspectral images with high fidelity even in the presence of significant noise.

\section{Method}
\label{sec:method}

\begin{table*}[h!]
    \centering
    \caption{\textbf{Comparison of SOTA hyperspectral imaging systems} in terms of reconstruction performance, and computational and optical requirements. Reconstruction quality is benchmarked on 30 images from the Harvard dataset~\cite{chakrabarti2011statistics} under a brightly lit condition with 5-second total exposure time and optical component effects (see supplement). Timings are reported for an NVIDIA RTX A6000. The count of optical components includes: lenses, apertures, prisms, actuators, SLMs, and sensors, but not control electronics.}
    \begin{adjustbox}{max width=\textwidth}
    \begin{tabular}{lcccccccc}
        \toprule
        \multirow{2}{*}{Modality} & \multicolumn{3}{c}{Reconstruction Performance} & \multicolumn{3}{c}{Computation and Optics} & \multirow{2}{*}{Category} \\
        \cmidrule(lr){2-4} \cmidrule(lr){5-7}
        & PSNR (dB) $\uparrow$ & SSIM $\uparrow$ & SAM ($^\circ$) $\downarrow$ & Compute Time & FoV ($^\circ$) & Opt. Components & \\
        \midrule
        Tunable Filter  & 4.78 & 0.08 & 70.10 & 0 & N.A. & 3  & Spectral Filter \\
        S.DiffuserCam \cite{monakhova2020spectral} & 15.59 & 0.46 & 34.98 & $>$ 15 mins & 14 & 3 & Spectral Filter \\
        S.DefocusCam \cite{foley2025spectral} & 21.17 & 0.61 & 25.11 & 39 s & 8 & 4 & Spectral Filter \\
        MST \cite{cai2022mask} & \underline{30.62} & \textbf{0.92} & 9.33 & 1.19 s & 10 & 9  & Dispersive Element \\
        Choi et al. \cite{choi2017high} & 23.97 & 0.68 & 14.85 & $>$ 15 mins & 10 & 9 & Dispersive Element \\
        KRISM \cite{saragadam2019krism} & 29.40 & 0.91 & \textbf{6.91} & 0 & 32 & 20 & Dispersive Element \\
        2in1 Cameras \cite{shi2024split} & 29.11 & 0.87 & 8.72 & 1.74 s & 9 & 4 & DOE\\
        Zhan et al. \cite{zhan2019hyperspectral} & 14.11 & 0.07 & 23.95 & 0.24 s & 10 & 5 & Chromatic Aberration \\
        \textbf{Ours} & \textbf{30.81} & \textbf{0.92} & \underline{7.35} & 0.64 s & 10 & 4 & Chromatic Aberration \\
        \bottomrule
    \end{tabular}
    \end{adjustbox}
    \label{tab:comparison}
\end{table*}

We aim to reconstruct a hyperspectral image $\bX \in \mathbb{R}^{H \cdot W \times C}$ of the target scene using a focal stack $\bY \in \mathbb{R}^{H \cdot W \times N}$, where $H$, $W$, and $C$ represent the height, width and channel of the image, respectively.
Since $N\ll C$ grayscale images in the focal stack are captured by translating the moving lens to different positions $z_i$ (more details in \cref{subsec:optical system}), we can express the forward model between $\bX$ and $\bY$ as follows:
\begin{align}
    \label{eq:forward_linear}
    \by = \bC \bH \bx = \bC
    \begin{bmatrix}
        \bH_{1,1} & \dots & \bH_{1,C} \\
        \vdots & \ddots & \vdots \\
        \bH_{N,1} & \dots & \bH_{N,C}
    \end{bmatrix}
    \bx.
\end{align}
Each block $\bH_{i,j} \in \mathbb{R}^{ (H+K-1) \cdot (W+K-1) \times H \cdot W}$ is a linear 2D convolutional matrix generated by a PSF $\bK(z_i, \lambda_j) \in \mathbb{R}^{K \times K}$, associated with the $j$th channel of $\bX$ and wavelength $\lambda_j$.
The binary matrix $\mathbf{C} \in \mathbb{R}^{N \cdot H \cdot W \times N \cdot (H+K-1) \cdot (W+K-1)}$ is used to crop the targeted region of the convolved result (i.e., $\bH \bx$), ensuring the same pixel count between $\bx$ and $\by$. 
We vectorize the images $\bX$ and $\bY$, defining $\bx = \operatorname{vec}(\bX) := [\bX_1^T, \dots, \bX_C^T]^T$ and $\by = \operatorname{vec}(\bY)$, where $\bX_j$ denotes the $j$th column of the matrix $\bX$.
Here, $\bX_j$ can also be interpreted as the $j$th channel of the hyperspectral image. Unless otherwise specified $N=5$ is assumed throughout (see supplement).

\subsection{System Design}
\label{subsec:optical system}

As shown in Fig.~\ref{fig:system}a, the imaging system consists of an objective lens, a movable lens, and a grayscale sensor.
The objective lens and the camera remain fixed while the second lens is translated to $N$ positions along the optical axis, $z_1,\cdots, z_N$, to capture the $N$ images, forming a chromatic focal stack. Each lens position $z_i$ is calibrated so that the corresponding wavelength $\lambda_i$ is in focus, as illustrated by the measured point spread functions (PSFs) in Fig.~\ref{fig:system}b. When the scene lies within the system’s depth of field, the captured image $I_i$ contains sharp structures at wavelength $\lambda_i$, while other wavelengths appear increasingly blurred as they deviate from $\lambda_i$.
The total travel distance of the moving lens, $z_N - z_1$, is small compared to the system’s effective focal length, so we assume the magnification of $I_1, \ldots, I_N$ remains constant across the stack.

\subsection{Inverse Algorithm}
\label{subsec:algorithm}

We solve the spectral image recovery problem with an iterative optimization that alternates between inverting the physical forward model in \cref{eq:forward_linear} and applying a data-driven regularizer. Specifically, we modify plug-and-play ADMM \cite{chan2016plug} with two task-specific augmentations to improve robustness and computational speed. First, we project spectra onto a limited number of principal components, taking advantage of the low-rank statistics of natural visible spectra. Second, we separate the spatial and frequency domain computations for fast matrix inversions. 

We formulate a minimization problem in an eigenspace as follows:
\begin{align} \label{eq:inverse_problem}
    \min_\bz \frac{1}{2} ||\by - \bC \bH \bP \bz||_2^2 + \Phi_{\boldsymbol{\theta}}(\bP \bz),
\end{align}
where $\Phi_{\boldsymbol{\theta}}(\cdot)$ is an off-the-shelf deep-learning-based regularizer with parameters $\boldsymbol{\theta}$ \cite{lai2023hybrid}.
The vector $\bz \in \mathbb{R}^{H \cdot W \cdot v}$, where $v$ represents the dimensionality of the selected eigenspace, is utilized to exploit the property that natural visible-range spectra primarily reside in a lower-dimensional space (i.e., $\bx \approx \bP \bz$).
We define $\bP=\bB^T \kron \bI_{HW}$, where the rows of $\bB$, shaped as $\mathbb{R}^{v \times C}$, are eigenvectors obtained from the Harvard dataset \cite{chakrabarti2011statistics}, as reported in~\cite{aydin2024hypercolorization}.
The matrix $\bI_{HW}$ and the operator $\kron$ denote an identity matrix shaped as $HW \times HW$ and the Kronecker product, respectively.

To adopt a plug-and-play ADMM, we transformed \cref{eq:inverse_problem} by introducing slack variables $\bv$ and $\bu$, and by using the change of variables $\hat{\bH} := \bH \bP$:
\begin{equation} \label{eq:inverse_problem_admm}
\begin{aligned}
    &\min_{\bz, \bu, \bv} \frac{1}{2}\| \by - \bC\bv \|_2^2 + \Phi_{\boldsymbol{\theta}} (\bu) \\
&\textrm{s.t.} \quad \bv = \hat{\bH} \bz,~ \bu=\bz.
\end{aligned}
\end{equation}
Followed by the derivation of ADMM (see supplement), we have \cref{alg:pnp_admm} for solving \cref{eq:inverse_problem_admm}.
Here, $\boldsymbol{\xi}$ and $\boldsymbol{\eta}$ denote dual variables.

\begin{algorithm} 
\caption{Plug-and-play ADMM Iterative Procedure} \label{alg:pnp_admm}
\begin{algorithmic}[1]
\State \textbf{Given:} $\mu_1, \mu_2 > 0$, and $\phi_{\boldsymbol{\theta}}(\cdot)$ as an off-the-shelf deep learning denoiser.
\State \textbf{Initialize:} $\mathbf{z}_0 \gets 0.5 \cdot \boldsymbol{1}$, $\mathbf{u}_0 \gets 0.5 \cdot \boldsymbol{1}$, $\boldsymbol{\xi}_0 \gets \boldsymbol{0}$, $\boldsymbol{\eta}_0 \gets \boldsymbol{0}$
\While{not converged}
    \State $ {\bv_{i+1} \leftarrow (\bC^T \bC + \mu_1 \bI)^{-1} (\bC^T \by + \mu_1 \hat{\bH} \bz_i - \boldsymbol{\xi}_i)}$
    \State $ {\bz_{i+1} \leftarrow (\mu_1 \hat{\bH}^T \hat{\bH} + \mu_2 \bI)^{-1} \Big(\hat{\bH}^T (\mu_1 \bv_i + \boldsymbol{\xi}_i)}$
    ${\hspace{5em} +  (\boldsymbol{\eta}_i + \mu_2 \bu_i)\Big)}$
    \State $ {\bu_{i+1} \leftarrow \bP^T \phi_{\boldsymbol{\theta}}\left(\bP(\bz_{i+1} + \boldsymbol{\eta}_i)\right)}$
    \State $ {\boldsymbol{\xi}_{i+1} \leftarrow \boldsymbol{\xi}_i + \mu_1 (\bv_{i+1} - \hat{\bH} \bz_{i+1})}$
    \State ${\boldsymbol{\eta}_{i+1} \leftarrow \boldsymbol{\eta}_i + \mu_2 (\bu_{i+1} - \bz_{i+1})}$
\EndWhile
\end{algorithmic}
\end{algorithm}

\begin{figure*}[ht!]
  \centering
  \includegraphics[width=\textwidth]{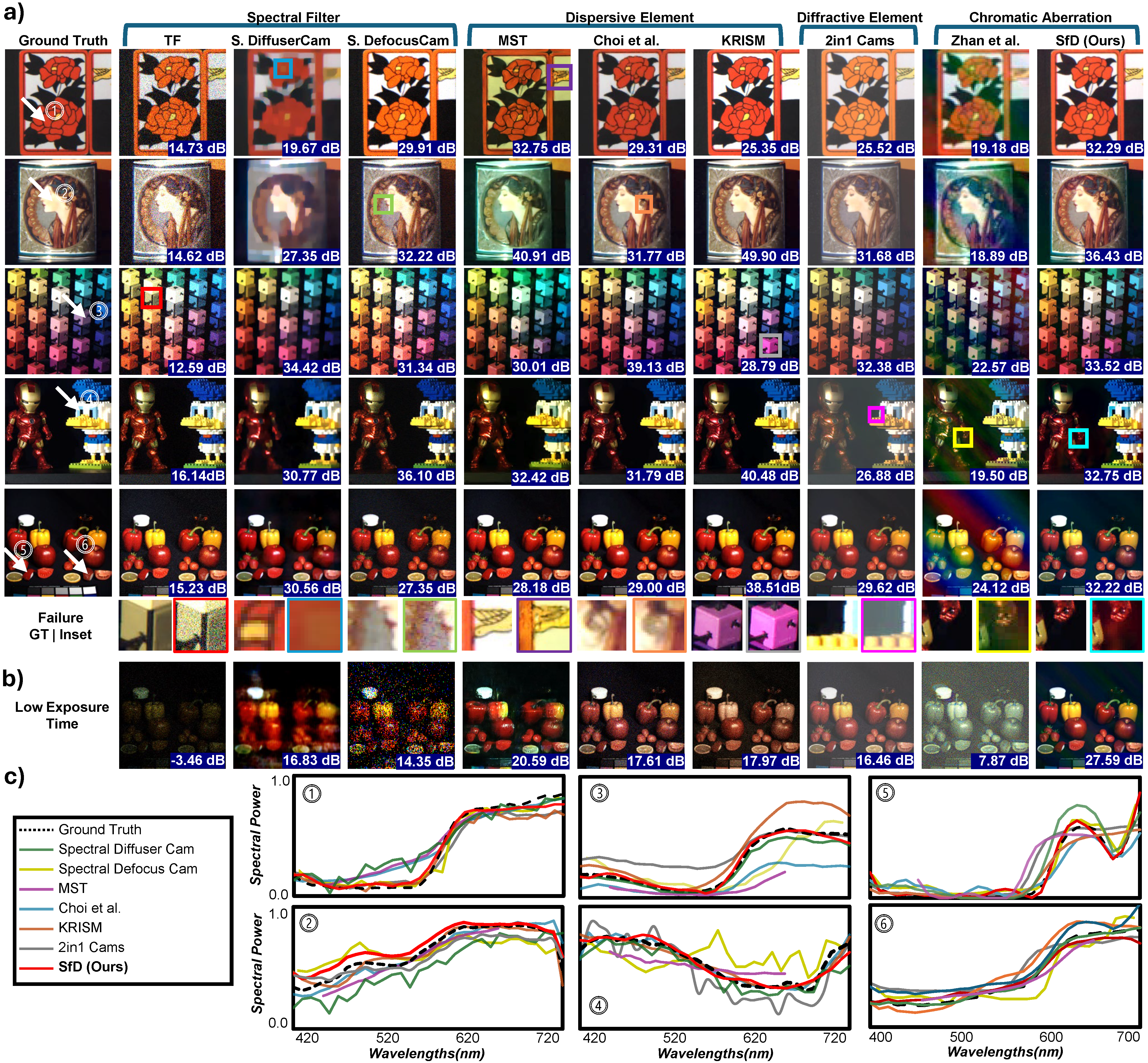}
    \caption{\textbf{Simulated reconstruction examples.}
    (a) Hyperspectral reconstructions from the KAIST~\cite{choi2017high} and CAVE~\cite{yasuma2010generalized} datasets, illustrated with RGB projections, annotated with hyperspectral PSNR. Insets highlight characteristic failures of each method, with comparison to ground truth. 
    (b) RGB projections and PSNR values under a low-exposure setting (2.9 s total exposure), where SfD has an even more significant advantage (see supplement).
    (c) Reconstructed spectra from the white arrow locations in (a). The selected points cover a range of colors, with points 5 and 6 illustrating the importance of spectral imaging with an RGB metamer. Note that MST and Spectral DefocusCam exhibit narrower spectral coverage, due to their training/calibration data and forward models, and that tunable filters and Zhan et al. are illustrated instead in supplement due to their high spectral noise.}
   \label{fig:sim_results} 
\end{figure*}

Plug-and-play ADMM primarily consists of two main update steps: a primal variable update ($\bv$, $\bz$, $\bu$) and a dual variable update ($\boldsymbol{\xi}$ and $\boldsymbol{\eta}$).
In the primal update, $\bv$ can be interpreted as a pseudo-measurement estimated from the true measurement $\by$ and the current update $\bz_i$.
An estimate of a projected image $z_{i+1}$ is then generated using a Wiener-like filter, $(\mu_1 \hat{\bH}^T \hat{\bH} + \mu_2 \bI)^{-1}$.
Since this estimate may contain noise, the proposed algorithm further applies an off-the-shelf deep learning denoiser \cite{lai2023hybrid} $\phi_{\boldsymbol{\theta}}(\cdot)$ to suppress noise and enhance the result.
See supplement for a comparison demonstrating the advantage of this deep denoiser over $\ell_1$ regularization.
We applied the denoiser after converting $\bz_{i+1}$ back to the image domain, as the off-the-shelf denoiser was trained on data in the hyperspectral image domain.
In the dual update, $\boldsymbol{\xi}$ and $\boldsymbol{\eta}$ are updated by evaluating the residual of the constraints in \cref{eq:inverse_problem_admm}. 
By iteratively alternating between primal and dual updates, the algorithm can heuristically converge to produce a projected image, which can then be used to recover a hyperspectral image (i.e., $\bx \approx \bP \bz$).
Specifically, we declare convergence when the step size falls below a fixed threshold or exceeds the previous step size scaled by a convergence check parameter.
It should be noted that updating $\mathbf{z}$ requires inverting a large, non-diagonal matrix.
Given the matrix's large size, computing its inverse directly is impractical.
To overcome this challenge, we derived a fast matrix inversion formula by leveraging the Block Circulant with Circulant Blocks (BCCB) structure of submatrices in $\bH$ (see supplement).
This fast inversion formula enables us to run the full algorithm with sub-second reconstruction times.
As a final step, we correct for the spectral response curve of our grayscale camera and visible light filter.
Additionally, see supplement for the settings of $\mu_1$ and $\mu_2$ in simulated and real data.

\section{Experimental Results}
We evaluate the performance of our method against several state-of-the-art methods in simulation on three publicly available hyperspectral datasets \cite{choi2017high, yasuma2010generalized, chakrabarti2011statistics}, then demonstrate its real-world performance with a hardware prototype.

\subsection{Simulation Settings}
All modalities are simulated for the same total exposure time (i.e., number of exposures~$\times$~duration per exposure) so that snapshot and multi-exposure methods have equivalent acquisition time and total photon budget.  We assume a brightly lit condition ($7.5 \cdot 10^{17}~\text{m}^{-2}\text{s}^{-1}$ in the visible range) and ideal cameras (negligible transition time between exposures and no quantization or readout noise). The simulations include Poisson noise and account for the light efficiency of optical components (see supplement).
Unless otherwise stated, the total exposure time is 5~seconds. We show in \cref{fig:sim_results}b and supplement that the advantages of our method are even stronger for lower total exposure times.

\subsection{Simulation Results}
As shown in \cref{tab:comparison} and illustrated in \cref{fig:sim_results}, SfD achieves the best PNSR and SSIM, and second-best SAM across 9 methods.
The performance of our method arises from its combination of photon-efficient optics and a physics-based reconstruction that remains stable under Poisson noise.

The spectral-filter-based modalities exhibit substantially lower spectral accuracy (see SAM in \cref{tab:comparison} and \cref{fig:sim_results}c1–4), primarily due to their lower photon efficiency (see supplement). The reduced photon efficiency leads to increased image noise or loss of high spatial frequency from denoising (see insets in \cref{fig:sim_results}), and the degradation becomes more pronounced under limited exposure conditions (see \cref{fig:sim_results}b).

In contrast, dispersive-based modalities somewhat mitigate photon loss at the cost of a more complex optical setup (see opt.~components in \cref{tab:comparison}). This optical complexity leads to relatively lower photon efficiency than SfD, causing haziness and other color errors, particularly when the photon budget is limited. Note that the purely data-driven MST method, despite its high performance in \cref{tab:comparison}, exhibits severe hallucinations in the low exposure time case (see yellow/red pepper in \cref{fig:sim_results}b). KRISM achieves the highest SAM performance by applying per-scene singular value decomposition (SVD). However, its per-scene basis can introduce spectral bias, as the SVD optimization may favor colors specific to each scene (see oversaturation in the inset and undersaturation in the low exposure image). 
The DOE-based 2in1 Camera loses significant photons to its RGB filters, and its data-driven model introduces haze and undersaturation for many scenes.

Chromatic-aberration-based modalities, including Zhan et al. and ours, offer greatly improved photon efficiency (see supplement). However, high photon efficiency does not guarantee robust reconstruction. In particular, Zhan et al.’s method relies on naive inverse filtering, which becomes highly unstable even under mild noise. 

Our method produces SOTA hyperspectral results, with robust performance across scenes from all datasets. Errors visible in RGB projections are rare, but see inset for color bleeding into dark regions. Small spectral errors can be seen in \cref{fig:sim_results}c, which shows that reconstructed spectra can be slightly oversmoothed by our use of a low-rank hyperspectral representation. Importantly, our SfD hardware preserves nearly all incident light and addresses shot noise witha robust reconstruction algorithm enables, for stable performance even under low exposure conditions. \cref{fig:sim_results}b shows a dramatic advantage to SfD in low light levels, demonstrating that our physics-based iterative reconstruction produces sharp, accurate results with minimal hallucination while still maintaining sub-second runtime.

\begin{figure*}[ht!]
  \centering
  \includegraphics[width=\textwidth]{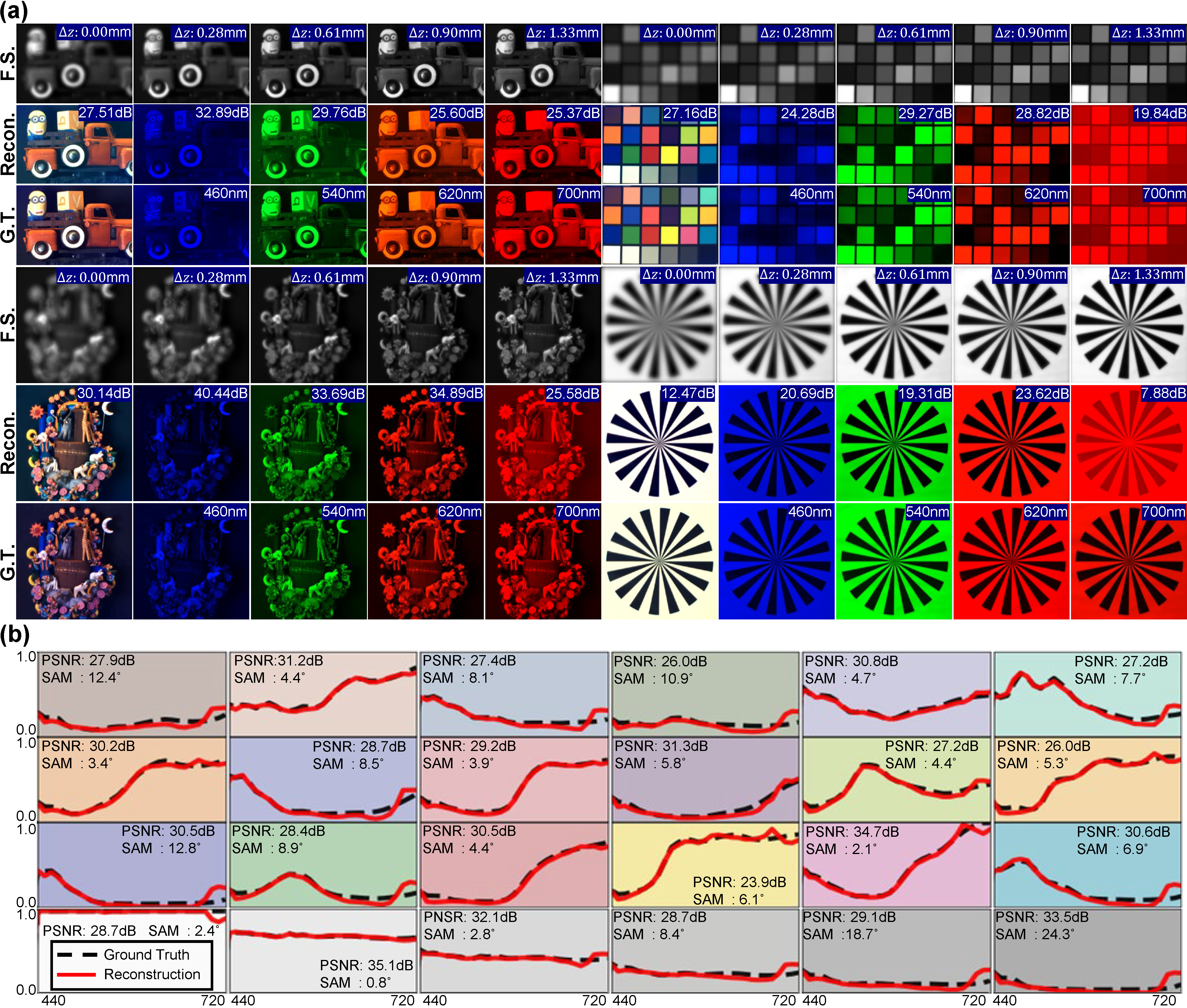}
   \caption{\textbf{Results from SfD prototype.} (a) We show raw grayscale measurements from the camera, samples from corresponding per-channel reconstruction results, and the aligned ground truth for four real scenes. $\Delta z$ indicates the lens displacement relative to the initial position, and sampled  channels have an approximate bandwidth of $10$nm centered at the displayed wavelength. The reconstructions demonstrate faithful recovery of both spatial details and spectral content. (b) Reconstructed and ground truth spectral curves for each Macbeth color patch. 
   }
   \label{fig:hw_results}
\end{figure*}

\subsection{Data Collection}
\label{subsec:system_specifications}
We built a prototype according to \cref{sec:method}.
Our prototype consists of a grayscale sensor and two 50 mm off-the-shelf lenses, where the focusing lens is mounted on a linear stage (see supplement for more details including a full parts list). The two selected lenses make longitudinal chromatic aberration the dominant aberration in our system (see supplement). Our optical design has a depth of field of 34 cm (2.64–2.98 m) and an axial chromatic focal shift of approximately 0.7 mm across the visible spectrum (see supplement), which we use as the primary source of spectral encoding.

We calibrated the system by imaging a point source at 2.8 m away from the camera. 
The source was a tungsten–halogen lamp covered by a $100$ $\mu$m pinhole, viewed through liquid crystal tunable filter (LCTFs).
The filter spanned $440-720$ nm in $10$ nm steps, providing 29 spectral bands. 
See supplement for the illuminant, filter, and sensor quantum efficiency spectra.
At each of five lens positions $z_i$, we measured point spread functions $\bK(z_i,\lambda_j)$ across all 29 bands (\cref{fig:system}b shows representative examples).
These measured PSFs are unrolled into our forward model $H$, and used during reconstruction. 

Ground-truth spectral images were captured using Liquid Crystal Tunable Filters (LCTFs), with the total exposure time slightly adjusted
to accommodate the scene’s dynamic range. A flat spectral target was used to correct per-channel filter absorption. For all data, we consider the central $800$ by $800$ pixels (9.7 deg FOV) cropped from the $1920$ by $1200$ image. All targets were placed 2.8 m from the camera unless otherwise noted, for minimal depth defocus. See supplement for details, all code and data on our Github page.

\subsection{Results on Real Data}
Our SfD camera produces sharp and accurate spectral reconstructions from blurry focal stacks and exhibits high performance consistent with simulations. \cref{fig:hw_results}a shows a subset of the collected data and per-channel ground truth, reconstruction, and PSNR for four real scenes. Observe that a variety of colorful objects are reproduced faithfully, both within each channel and in their collective color recovery. Results on all channels for these scenes and the book in \cref{fig:teaser} are shown in
supplement. For comparison to a direct measurement baseline, where chromatically-aberrated images focused at 671, 546, and 442 nm are used as RGB channels, see supplement.

Note that, while nearly all channels on nearly all scenes produce high-quality reconstructions, some channels of the pinwheel target (see 700 nm in \cref{fig:hw_results}a) show a failure to reconstruct the full resolution and contrast of the pattern. This is likely due to the difficulty in  distinguishing between true black regions that have been brightened by defocused light from neighboring white pixels, versus gray regions. This failure case could be mitigated with a sensor of higher dynamic range, using more than five measurements, or by incorporating priors that favor target-like objects over natural scene statistics (e.g. total variation in place of or in addition to the deep denoiser). 

\cref{fig:hw_results}b shows spectral reconstruction accuracy for all Macbeth ColorChecker patches. The results show an average PSNR of $29.54$ dB and a SAM of $7.42^\circ$ (compare to 30.81 dB and 7.35$^\circ$ in simulation from \cref{tab:comparison}). Note that color checkers provide challenging input to our method due to their lack of natural texture, and that spectral errors are higher at longer wavelengths, where our camera's focal shift is less informative (see supplement). This performance trend highlights the underlying working principle of SfD: stronger chromatic focal shifts yield better spectral separation, which in turn improves reconstruction accuracy.

\begin{figure}[ht!]
  \centering
  \includegraphics[width=\columnwidth]{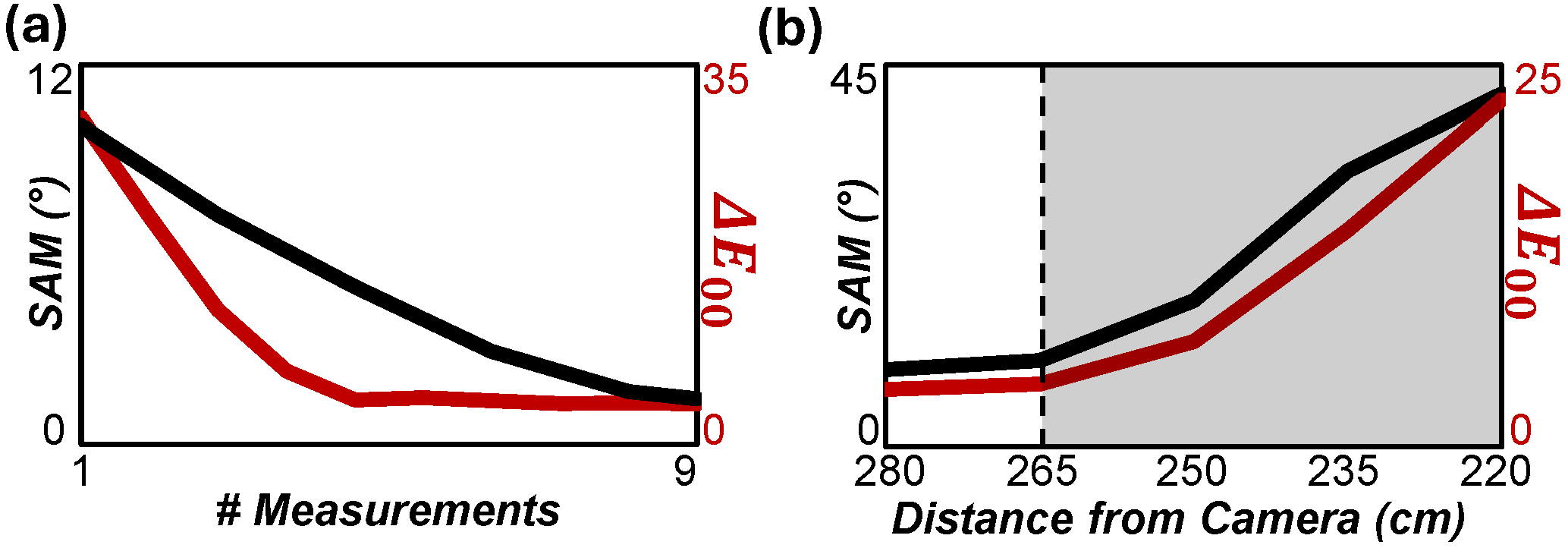}
   \caption{\textbf{Robust Recovery.} 
    (a) Adding measurements improves reconstruction, with performance saturating more quickly in RGB than in spectrum. 
    $\Delta E_{00}$ is a standard color perception metric \cite{sharma2005ciede2000}, which we adopt here to evaluate RGB reconstruction quality.
    (b) SfD performs robustly within and somewhat beyond the predicted working range of 264-298 cm (dashed vertical line).
    }
   \label{fig:trends}
\end{figure}

We show in \cref{fig:trends} the robustness of SfD on RGB and spectral color recovery using the color checker. \cref{fig:trends}a demonstrates that additional measurements stabilize reconstruction, with an effect that naturally saturates earlier on the lower-dimensional RGB reconstruction (see supplement for corresponding images and spectra).
The RGB and hyperspectral reconstruction quality remains high within the expected depth range (white region in \cref{fig:trends}b). Outside this range, the reconstruction quality decreases steadily. The supplement shows that chromatic low-frequency artifacts only begin to appear when the object depth approaches 2.35 m—approximately 29 cm beyond the theoretical range. This demonstrates the robustness of SfD in practice, even when the working range assumption is mildly violated.

\section{Conclusion}

We have demonstrated Spectrum from Defocus (SfD), which reconstructs hyperspectral images from chromatically aberrated focal stacks.  
Our camera, consisting of two off-the-shelf lenses and a grayscale sensor, achieves SOTA performance and fast reconstruction with an interpretable physics- and data-based iterative algorithm. 
We demonstrate its performance in simulation against eight other methods, with functional principles including spectral filters, dispersive elements, diffractive optical elements, and chromatic focal stacking, apried with reconstruction algorithms based on physical models, data-driven networks, and hybrid approaches. Our hardware prototype matches simulation performance on real-world data.

Our proof-of-concept prototype also reveals several limitations.
First, we assume scenes occupy a working range of 34 cm (see supplement, but note robustness beyond theoretical predictions demonstrated in \cref{fig:trends}) and a narrow paraxial field of view (see supplement, but note comparable limitations on almost all SOTA methods in \cref{tab:comparison}).
Expanding the working volume will require joint optical and algorithmic design to ensure that signals remain strong and recoverable across a broader range of conditions.
Second, red and black colors experience slightly lower reconstruction accuracy. This is because our chromatic focal shift
flattens at longer wavelengths, which could be improved with a more dispersive optical glass, 
and because the system exhibits limited dynamic range. Rare failure cases tend to occur in black regions, which can appear gray or falsely colored due to sensor saturation and light bleed from neighboring bright pixels.
Third, because the reconstruction must model incoming photons of all wavelengths, a spectral reconstruction is required in order to produce RGB measurements. In practice, this means that five measurements are required to produce high-quality RGB images. 
Reducing the effects of multi-shot imaging, using tools from burst photography \cite{dudhane2024burst,hasinoff2016burst,lee2025ntire} and greater reliance on data-driven reconstruction, is an important direction for SfD in dynamic scenes.

Finally, while the light efficiency and robust reconstruction of our method are well-suited to fast hyperspectral imaging, and preliminary results on low exposure times suggests a significant advantage to our method in photon-starved settings (see \cref{fig:sim_results}b and supplement), 
exploring this advantage will require higher fidelity noise models for low-light images~\cite{monakhova2022dancing,lu2025dark}.

\FloatBarrier
{
    \small
    \bibliographystyle{ieeenat_fullname}
    \bibliography{main}
}


\end{document}